\documentclass[twoside]{article}

\usepackage[preprint]{aistats2026}

\usepackage{graphicx}
\usepackage{subcaption}
\usepackage{booktabs}

\usepackage{algorithm}
\usepackage{algorithmic}

\usepackage{amsmath}
\usepackage{amssymb}
\usepackage{amsthm}
\usepackage{xcolor}
\usepackage{microtype}

\usepackage[round,authoryear]{natbib}

\usepackage[hidelinks]{hyperref}

\input{Definitions}

\begin{document}

\twocolumn[

\aistatstitle{Adaptive Data Harvesting for Efficient Neural Network Learning with Universal Constraints}

\aistatsauthor{
Siteng Kang \And Xinhua Zhang
}

\aistatsaddress{
University of Illinois Chicago \\
\texttt{\{skang98, zhangx\}@uic.edu}
}
\begin{center}
\textit{Preprint}
\end{center}
]

\begin{abstract}
Training neural networks to satisfy universal constraints over continuous domains poses unique challenges.
Common examples include Lyapunov Neural Networks (Lyapunov NNs) and Physics-Informed Neural Networks (PINNs),
where analytical solutions are generally either unavailable or overly restrictive.
Sample-based methods are therefore commonly used to enforce these constraints, 
and the choice of samples has a substantial impact on convergence speed, stability, and solution quality.
Most existing methods rely on fixed heuristics or handcrafted rules,
and are suboptimal in practice.
In this paper, we aim to improve upon them by learning---from data and experience---how to dynamically and iteratively adjust the samples in response to the model’s evolving learning performance.
Trained by reinforcement learning,
the learned policy improves empirical constraint satisfaction on test problems while significantly improving efficiency.
We validate the approach on both Lyapunov NNs and PINNs, 
and demonstrate its broader applicability to domains where adaptive input selection is essential for effective training.
\end{abstract}
\section{Introduction}
\label{adh:introduction}

Many neural network applications are governed by universal constraints that must be satisfied across continuous input domains.
Two prominent examples are Lyapunov Neural Networks (Lyapunov NNs), 
which learn Lyapunov functions to certify the stability of nonlinear dynamical systems, 
and Physics-Informed Neural Networks~\citep[PINNs]{raissi2019physics},
which enforce physical laws encoded as partial differential equations (PDEs).
A defining challenge in these settings is that the underlying constraints%
---such as the Lyapunov decrease condition or PDE residuals in PINNs---%
must hold throughout the entire input space, not just at isolated points. 
Since the input space is infinite, exhaustive verification of universal constraints is infeasible.
As a result, training relies on selecting a representative set of collocation points to guide optimization.
The quality, distribution, and adaptivity of these points directly influence the model's ability to generalize—that is, to satisfy the constraints at unseen locations. 
Poorly chosen samples may lead to localized overfitting and instability, 
while effectively chosen samples improve training efficiency, convergence speed, and overall solution fidelity.

Lyapunov Neural Networks aim to certify stability by ensuring the Lyapunov condition holds throughout a continuous state space.
Early work by \cite{richards2018lyapunov} proposed an iterative scheme that expands level sets and samples new points from the expanded safe regions. 
Subsequent methods \citep{chang2019neural,dai2021lyapunov} introduced counterexample-guided updates to improve training efficiency.
\cite{abate2020formal} formalized this approach with theoretical guarantees,
while \cite{wu2023neural} incorporated a buffer mechanism to retain and reuse previously identified counterexamples. 
Despite these advances, existing methods still rely on manually tuned expansion rules or computationally expensive verification procedures, 
which limit the adaptability and scalability in practice.

A similar challenge arises in the PINN literature. 
Early methods~\citep{raissi2019physics} adopted fixed sampling strategies such as uniform grids or random draws.
While simple to implement, these methods may under-represent regions of interest or yield uneven spatial coverage. 
Quasi-random schemes such as \cite{Sobol1967OnTD} or \cite{Halton1960OnTE} sequences offer better coverage but remain static and model-agnostic. 
To improve performance, adaptive techniques like Residual-based Adaptive Sampling (RAD)~\citep{wu2023comprehensive} have been proposed, 
where sampling density is adjusted based on local residual magnitudes. 
Other works have explored adversarial sampling~\citep{tang2024adversarial}, 
importance sampling based on residuals or gradients~\citep{li2025importance}, 
and multi-stage training schemes~\citep{wang2024multistage}.
While these heuristics improve local sample efficiency, 
they typically rely on \textbf{fixed rules or handcrafted criteria} and are not designed to adapt continuously to the evolving training dynamics. 

\textbf{In this work}, we propose a general framework that adaptively integrates heuristic sampling methods through a policy-driven selection mechanism. 
To this end, we formulate the process of selecting multiple batches of training points as a sequential decision-making problem, 
where a learned policy iteratively chooses samples from a dynamic candidate pool based on the feedback from the current model state. 
This design enables the selection strategy to evolve over time, 
continuously adjusting to training dynamics and model uncertainty. 
Unlike static heuristics or fixed schedules, our approach is model-agnostic, 
supports training with standard reinforcement learning (RL) algorithms such as TD3~\citep{fujimoto2018td3} and SAC~\citep{haarnoja2018sac}, 
and requires no task-specific modifications.

We evaluate this framework on both Lyapunov NNs and PINNs, 
showing consistent improvements in convergence speed and final accuracy compared to baselines.
Our results highlight the benefit of coupling adaptive sampling with policy-driven selection in universally constrained neural networks and suggest broader applicability to other data-centric training pipelines.
\section{Preliminary}
\label{adh:preliminary}
The proposed framework can be applied to a range of domains, 
including Lyapunov-based stability certification,
Physics-Informed Neural Networks,
nonnegative function learning~\citep{kim2024inverse},
Lipschitz continuity~\citep{virmaux2018lipschitz}, etc.
This work focuses on the first two domains,
leaving the rest to future work.
We will first contextualize our approach by reviewing relevant prior methods in these areas. 

\subsection{Lyapunov Neural Network}
We begin by examining Lyapunov Neural Networks, 
where effective input sampling strategies are critical for robustly certifying stability in control systems.
Lyapunov NNs are neural-network-based representations of Lyapunov functions
—a scalar, positive definite function that decreases along the trajectories of a dynamical system. 
Such a function certifies that the system state will asymptotically converge to an equilibrium (w.l.o.g., the origin).
Specifically, we consider a deterministic, discrete-time dynamical system:
\begin{align}
x_{t+1} = f^{\psi}(x_t),
\end{align}
where $x_t \in \mathbb{X}$ represents the system state, $t\in \mathbb{N}$ is the time step, and $\psi$ is a fixed policy. 
Let $S^{\psi}$ denote the true region of attraction (ROA):
the set of all states that converge to zero and remain in a bounded neighborhood of $0$ under the policy $\psi$.
The goal is to learn a Lyapunov function $v_{\theta}(x)$, 
parameterized by a neural network with weights $\theta$, 
that certifies as large a subset of $S^{\psi}$ to be safe as possible. 
According to Lyapunov’s Stability Theorem~\cite{kalman1960control}, if $v_{\theta}$ is positive definite and
$\Delta v_{\theta}(x) := v_{\theta}(f^{\psi}(x)) - v_{\theta}(x) < 0$ \textbf{for all} $x \ne 0$ such that $v_{\theta}(x) \le c$, 
then the sublevel set $V_{\theta}(c) = \{x : v_{\theta}(x) \le c\}$ is an invariant safe set,
i.e., $V_{\theta}(c) \subseteq S^{\psi}$.

Training a Lyapunov NN is a non-standard learning problem: instead of minimizing prediction error, 
the goal is to satisfy inequality constraints throughout a continuous state space. 
Since exact verification is intractable, 
sampling becomes necessary. 
A well-chosen set of sample states must sufficiently cover both the core and the boundary of the ROA,
particularly the regions where the Lyapunov condition is hardest to satisfy.

\begin{algorithm}[t]
\caption{ROA Classifier Training~\citep{richards2018lyapunov}}
\label{alg:roa-original}
\begin{algorithmic}[1]
\REQUIRE closed-loop dynamics $f^{\psi}$, expansion multiplier $\alpha > 1$, simulation horizon $T$
\ENSURE{Lyapunov NN $v_\theta$ for testing dynamics}
\STATE Initialize Lyapunov NN $v_{\theta}$ and safety level $c_0$.
\FOR{$k$ = 0 to \# ROA\_resample}
    \STATE Sample a subset from $V_{\theta}(\alpha c_k)$, and denote it as $X$.
    \STATE Forward simulate the batch for $T$ steps: $S \gets \{x \in X  : f^{\psi}_{(T)}(x) \in V_{\theta}(c_k) \}$.
    \STATE Train $v_{\theta}$ on $X$ using hinge loss,
    with $S$ as the positive example set,
    and $X \backslash S$ the negative set.
    \STATE $c_{k+1} \gets \max_{x\in S} v_{\theta}(x)$.
\ENDFOR
\end{algorithmic}
\end{algorithm}

\cite{richards2018lyapunov} addressed this challenge via an iterative algorithm that formulates ROA estimation as a binary classification problem. 
At each iteration, they evaluate the current Lyapunov candidate $v_{\theta}$ on a batch of sampled states, 
labeling them as safe or unsafe depending on whether their simulated trajectories satisfy the Lyapunov condition. 
Then, they expand the safe level set $V_{\theta}(c_k)$ by a fixed expansion multiplier $\alpha > 1$,
and draw new samples from this expanded region. 
A classification loss is used to update $\theta$, and the process repeats.
Algorithm~\ref{alg:roa-original} shows their training procedure.

In practice, however, the selection of $\alpha$ can be highly delicate,
requiring adaptation across the training process.
An overly conservative value leads to slow progress,
while an overly aggressive value generates unsafe samples.
Therefore, our framework will formulate the process as a sequential decision-making problem.
We train an RL agent to dynamically select the expansion multiplier $\alpha$, 
enabling the sampling strategy to adapt continuously based on the evolving model state and training feedback. 
This removes the need for manual heuristics or external falsifiers and promotes more efficient and robust training.

\subsection{Physics-Informed Neural Network}
We now turn to Physics-Informed Neural Networks, 
another domain where effective sampling of input points significantly impacts training efficiency and accuracy.
Partial differential equations (PDEs) are mathematical equations that describe how physical quantities such as heat, fluid velocity, or electromagnetic fields evolve over space and time. 
Despite their ubiquity in modeling real-world phenomena, solving PDEs is notoriously difficult,
especially in high dimensions or with complex boundary conditions,
due to the need for fine-grained discretizations and numerical stability.

Physics-Informed Neural Networks~\citep[PINNs]{raissi2019physics} are deep learning models designed to solve PDEs by embedding the underlying physical laws directly into the learning objective. 
Unlike traditional numerical methods, PINNs approximate the solution $u(\textbf{x})$ with a neural network $u_\theta(\textbf{x})$, trained by minimizing the residual of PDEs at \textit{selected} points, known as \textit{collocation points}. 

We consider the general form of a forward PDE problem parameterized by $\lambda$ on a domain $\Omega\in\mathrm{I\!R}^d$, 
\begin{align}
\label{eq:pde}
f(\textbf{x};u, \lambda)
:= f \rbr{\textbf{x};\frac{\partial u}{\partial x_1},
\cdots, \frac{\partial^2 u}{\partial x_1\partial x_d},
\cdots;\textbf{$\lambda$}}= 0,
\end{align}
where the output of $f$ is a vector to encode multiple equalities.
Here the equality needs to be satisfied \textbf{universally} for all $\textbf{x} \in \Omega$.
Higher order partial derivatives are also admissible.
Let the boundary conditions be $\Bcal(u,\textbf{x})=0$.
PINN is trained by using a loss function over a set of collocation points $X=X_f\bigcup X_b$, 
where $X_f$ is sampled inside the domain and $X_b$ is sampled on the boundaries: 
\begin{align}
\label{eq:obj_PINN}
    \Lcal_\lambda(\theta;X):= 
    \sum_{\textbf{x}\in X_f} 
    \nbr{ f(\textbf{x};u_\theta, \lambda)}^2
    + w\sum_{\textbf{x} \in X_b} \Bcal(u_\theta,\textbf{x})^2.
\end{align}

Our work introduces a reinforcement learning framework that formulates collocation selection as a sequential decision-making problem. 
An RL agent dynamically selects training points from an evolving candidate set at each iteration, 
using the feedback from the current model to optimize for long-term training efficiency. 
The framework is model-agnostic, 
compatible with standard RL algorithms such as TD3 and SAC, 
and continuously adapts to the changing training landscape. 
We will empirically show that it consistently outperforms both static baselines (uniform, random, Sobol, Halton) and adaptive heuristics such as RAD in convergence speed and final solution accuracy.

\paragraph{Parameterization of the sampler.}
In the most general sense,
a sampler is meant to draw upon the entire space where the universal constraint concerns.
To reduce variance,
\cite{richards2018lyapunov} effectively parameterized the sampling distribution using $\alpha c_k$ (step 3 of Algorithm~\ref{alg:roa-original}).
We will adopt the same approach and task the RL agent to propose $\alpha$.
Similarly, instead of directly parameterizing a distribution on the collocation space of PINN,
we resort to a mixture of expert model that leverages the samplers from existing literature,
and task the RL agent to propose their mixture weights.
These designs allow considerable flexibility for problem-specific customization.

\section{Adaptive Data Harvesting Framework}
\label{adh:framework}
Our  framework consists of two interacting components: 
an outer RL agent that guides the training of an evolving inner optimization entity (e.g., Lyapunov NN or a PINN). 
We formulate the outer RL component as a Markov Decision Process (MDP), 
defined by $\mathbb{M} = (\mathbb{S}, \mathbb{A}, \mathbb{P}, \mathbb{R}, \gamma, \mu_0)$, 
where $\mathbb{S}$ denotes the state space representing the current state of the inner optimization entity,
and $\mathbb{A}$ is the action space encompassing all possible sampled batches of training examples (or their distribution parameterization). 
The transition function $\mathbb{P}: \mathbb{S} \times \mathbb{A} \times \mathbb{S} \rightarrow [0,1]$ models
how the inner optimization entity evolves in response to selected actions,
while the reward function $\mathbb{R}: \mathbb{S} \times \mathbb{A} \rightarrow \mathrm{I\!R}$ quantifies the effectiveness of the selected examples. 
The discount factor $\gamma \in [0,1)$ controls the trade-off between immediate and future rewards, 
and $\mu_0: \mathbb{S} \rightarrow \mathrm{I\!R}$ specifies the initial state distribution. 

We next detail how to apply this RL-based framework to train Lyapunov NNs and PINNs.


\begin{algorithm}[t]
\caption{RL-Guided Adaptive Expansion ROA}
\label{alg:roa-rl}
\begin{algorithmic}[1]
\REQUIRE Testing dynamics $f^{\psi_\text{test}}$, simulation horizon $T$, initialization method $\mathcal{D}$
\ENSURE{Lyapunov NN $v_\theta$ for testing dynamics}
\STATE Initialize RL policy $\pi$.\hspace{6.5em} {\color{gray} // Training}
\FOR{$t$ = 1 to RL\_episode}
    \STATE Initialize random training dynamics $f^\psi$, Lyapunov NN $v_{\theta}$, safe level $c_0$.
    
        \FOR{k = 1 to ROA\_resample}
        \STATE Generate multiplier based on training state:\\ $\alpha_k = \pi(s_k)$.
        
        \STATE Sample a batch of points from the expanded region $V_{\theta}(\alpha_k c_k)$, and denote it as $X$.
        
        \STATE Forward simulate the batch for $T$ steps: \\ $S \gets \{x \in X  : f^{\psi}_{T}(x) \in V_{\theta}(c_k) \}$.
        
        \STATE Train $v_{\theta}$ on $X$ using labels $y_i = +1$ for $x_i \in S$, $y_i = -1$ otherwise.
        
        \STATE $c_{k+1} \gets \max_{x\in S} v_{\theta}(x)$, 
        compute $r_k$, update $\pi$.
    \ENDFOR
\ENDFOR
\STATE Initialize Lyapunov NN $v_{\theta}$, safe level $c_0$.{\color{gray} // Testing}
\FOR{$k$ = 1 to ROA\_resample}
    \STATE Generate multiplier based on training state $\alpha_k = \pi(s_k)$.
    \STATE Sample a batch of points from the expanded region $V_{\theta}(\alpha_k c_k)$, and denote it as $X$.
    \STATE Forward simulate the batch for $T$ steps: $S \gets \{x \in X : f^{\psi_\text{test}}_{T}(x) \in V_{\theta}(c_k) \}$.    
    \STATE Train $v_{\theta}$ on $X$ using labels $y_i = +1$ for $x_i \in S$, $y_i = -1$ otherwise.
    \STATE $c_{k+1} \gets \max_{x\in S} v_{\theta}(x)$.
\ENDFOR
\end{algorithmic}
\end{algorithm}

\subsection{Application to Lyapunov function learning}

We next show how adaptive data harvesting can be applied to  learning a Lyapunov function $v_{\theta}(x)$ with parameter $\theta$, 
certifying a large subset of $S^{\psi}$ as safe. 
To this end, we introduce an RL agent that adaptively selects the expansion multiplier $\alpha_k$ at each iteration based on the current state of training. 
As shown in Algorithm~\ref{alg:roa-rl}, the RL policy $\pi$ observes feedback $s_k$%
---such as the ratio of safe to unsafe samples---and outputs an expansion factor $\alpha_k = \pi(s_k)$. 
After each update to the Lyapunov network, a reward $r_k$ is computed based on the estimated ROA volume, 
and the policy $\pi$ is updated accordingly.

This approach enables dynamic control over the data sampling radius, 
leading to more adaptive and efficient training. 
Our experiments show that this adaptive strategy accelerates convergence and results in larger certified ROA regions compared to baselines with a fixed value of $\alpha$ .

\subsection{Application to PINN}

\begin{algorithm}[t!]
    \caption{RL-based collocation selector for PINN}
    \label{alg:pinn-rl}
    \begin{algorithmic}[1]    
        \REQUIRE{$n$ samplers $\{\mathcal{S}_{1:n}\}$}
        \ENSURE{trained RL policy $\pi$}
        \STATE  Initialize RL policy $\pi$. \hspace{6.5em} {\color{gray} // Training}
        \FOR{$t$ = 1 to \# RL\_episode}
            \STATE   Initialize a PINN $u_\theta$ with parameter $\theta$, and randomly sample a PDE with parameter $\lambda$.
            \FOR{$i$ = 1 to PINN\_resample}
                \FOR{$j$ in 1 to $n$ (\# base samplers)}
                    \STATE Sample collocations $X_j$ with sampler $\mathcal{S}_j(u_\theta)$
                    \STATE Compute the PDE Residual $\text{res}[j]$ for $X_j$ as $\Lcal_\lambda(u_\theta; X_j)$.                    
                \ENDFOR
                \STATE Generate sampler ratio: \\ $\mathbf{a} = (a_1, a_2, \dots, a_n)\sim \pi(\mathbf{a}| s = \text{res}[1:n])$.
                \STATE Create collocation set $X_\pi$ by drawing collocations from $X_{1:n}$ according to $\mathbf{a}$.
                \STATE Update $\theta$ according to $\Lcal_\lambda(u_\theta; X_\pi)$ for a pre-specified number of steps.
                \STATE Randomly generate test collocations $X_{\text{random}}$.
                \STATE $r \leftarrow$ negative error of $u_\theta$ on $X_{\text{random}}$,\\ and update $\pi$ by an off-line RL method.
            \ENDFOR
        \ENDFOR
        \STATE Initialize a PINN $u_\theta$ with parameter $\theta$.{\color{gray} // Testing}
        \FOR{i = 1 to PINN\_resample}
            \FOR{$j$ in 1 to $n$ (\# base sampler)}
                \STATE Sample collocations $X_j$ using sampler $\mathcal{S}_j(u_\theta)$.
                \STATE Compute the PDE Residual $\text{res}[j]$ for $X_j$ as $\Lcal_{{\color{red}\lambda_{\text{test}}}}(u_\theta; X_j)$.
            \ENDFOR
            \STATE Generate sampler ratio: \\ $\mathbf{a} = (a_1, a_2, \dots, a_n)\sim \pi(\mathbf{a}|s = \text{res}[1:n])$.
            \STATE Create collocation set $X_\pi$ by drawing collocations from $X_{1:n}$ according to $\mathbf{a}$.
            \STATE Update $\theta$ according to $\Lcal_{{\color{red}\lambda_{\text{test}}}}(u_\theta; X_\pi)$ for a pre-specified number of steps.
        \ENDFOR
    \end{algorithmic}
\end{algorithm}

We summarize in Algorithm~\ref{alg:pinn-rl} the full training and testing procedure for the outer RL-based collocation sampler $\pi$
and the inner optimization entity (a PINN).
At each RL training episode,
a new PDE is sampled with parameter $\lambda$,
and we initialize a new PINN with parameter $\theta$. 
During each collocation resampling step, we generate multiple candidate collocation sets using a collection of $n$ base samplers (e.g., those reviewed in Section~\ref{adh:introduction}),
and evaluate the PDE residual on each set. 
The resulting residuals collectively form the state provided to the RL agent,
which uses its policy $\pi$ to produce a set of mixing weights,
representing the sampling ratios across the baseline samplers. 
\begin{figure*}[t]
    \centering
    \begin{subfigure}[b]{0.45\textwidth}
        \centering
        \includegraphics[width=0.8\textwidth]{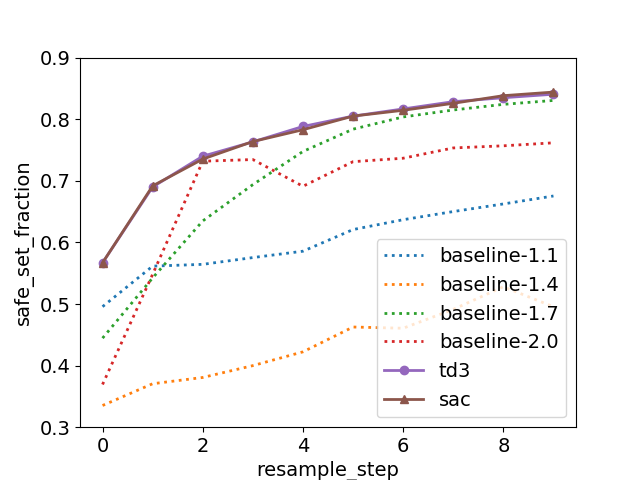}
        \caption{Testing safe set fraction after RL agent training}
        \label{fig:lyapunov-ssf-after}
    \end{subfigure}
    \begin{subfigure}[b]{0.45\textwidth}
        \centering
        \includegraphics[width=0.8\textwidth]{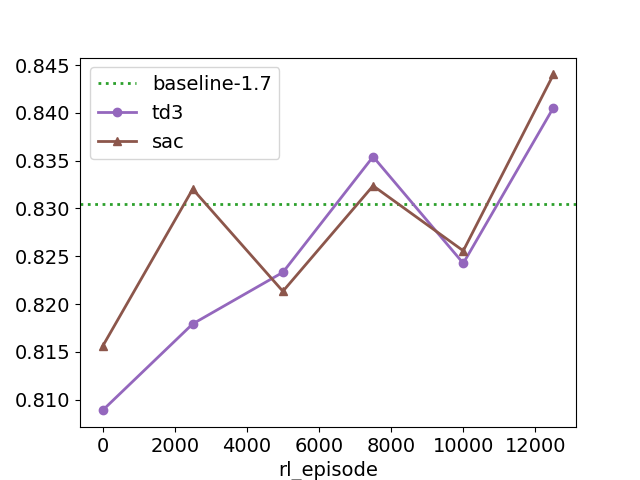}
        \caption{Testing safe set fraction after resampling step 10}
        \label{fig:lyapunov-ssf-step}
    \end{subfigure}
    \caption{Testing safe set fraction on Lyapunov-NN compared with baseline $\alpha$}%
    \label{fig:lyapunov-ssf}%
\end{figure*}
Collocation points are then drawn from the baseline samplers according to the selected ratio, 
and are used to update the PINN parameters $\theta$ by optimizing the objective in \eqref{eq:obj_PINN}. 
Afterwards, the performance of the resulting PINN is evaluated over a newly sampled set of collocations $X_{\text{random}}$.
Here, we use the error, averaged over $X_{\text{random}}$, comparing the learned PINN $u_\theta$ with the (near) ground-truth PINN $u$, 
which can be obtained either by closed form,
or by running any standard PINN solver to high accuracy.
The negative error serves as the reward signal to update the policy $\pi$ by any RL method such as SAC.

During testing, we apply the learned policy $\pi$ to a held-out PDE with parameter $\lambda_{\text{test}}$.
At each resampling iteration, the RL agent observes the residuals from candidate collocation sets and selects a sampling ratio, which is used to construct the collocation set for PINN training.

\section{Experimental Results}
\label{adh:experiment}
In this section, we empirically evaluate the effectiveness of our proposed framework for adaptive sample selection. 
We conduct experiments on four benchmark environments: 
Lyapunov Neural Network (on the Inverted Pendulum benchmark~\citep{richards2018lyapunov}),
PINN-Diffusion, PINN-Wave, and PINN-Burgers (from the PINN-sampling benchmark~\citep{wu2023comprehensive}).
Following standard evaluation protocols, each experiment is repeated with 5 random seeds, 
and we report the mean performance across trials.

For the Lyapunov neural network training, we evaluate our approach on the classic inverted pendulum benchmark, 
following the setup described by \cite{richards2018lyapunov}. 
The environment models a single-link pendulum actuated by a torque-limited motor, with dynamics governed by 
\begin{align}
    ml^2\ddot{\phi}=mgl\sin\phi-\beta\dot{\phi}+\tau,
\end{align}
where $\phi$ is the angular displacement from the upright equilibrium, $\tau$ is the input torque, $m$ is the pendulum mass, $g$ is the gravitational acceleration, $l$ is the pole length, and $\beta$ is the friction coefficient.
The continuous-time dynamics are discretized using a fixed time step of 0.01 seconds. 
The control input is subject to saturation, $\tau\in[-\bar{\tau},\bar{\tau}]$, reflecting physical actuator limits; 
as a result, the pendulum cannot be recovered once it strays too far from the vertical position. 
To stabilize the system, a linear quadratic regulator is applied around the upright equilibrium, 
yielding a local closed-loop ROA.
Within this ROA, the controller ensures convergence to the equilibrium, 
but outside it, the limited actuation cannot compensate for the nonlinear dynamics, 
the pendulum falls down, and the system trajectories inevitably diverge from the equilibrium. 
Training a Lyapunov certificate is especially difficult at the edge of the true region of attraction, 
since states can lie arbitrarily close to the safe/unsafe boundary and small classification errors can flip them between recovery and failure.
\subsection{Lyapunov Neural Network}

\begin{figure*}[t]
    \begin{minipage}{\textwidth}
    \centering
    \begin{subfigure}[b]{0.45\textwidth}
        \centering
        \includegraphics[width=0.8\textwidth]{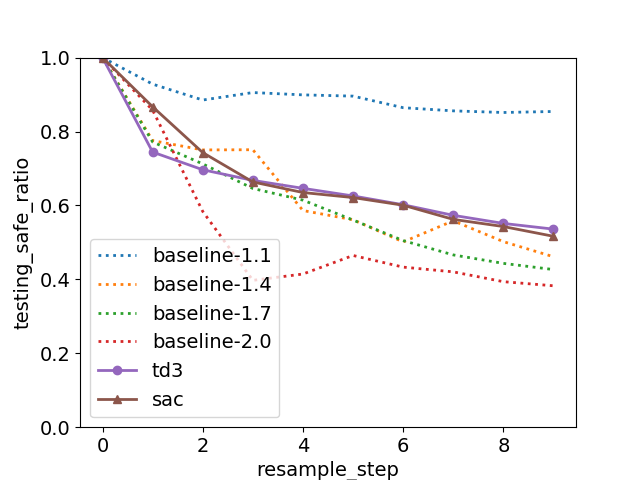}
        \caption{Testing safe ratio before RL agent training }
        \label{fig:lyapunov-gp-before}
    \end{subfigure}
    \begin{subfigure}[b]{0.45\textwidth}
        \centering
        \includegraphics[width=0.8\textwidth]{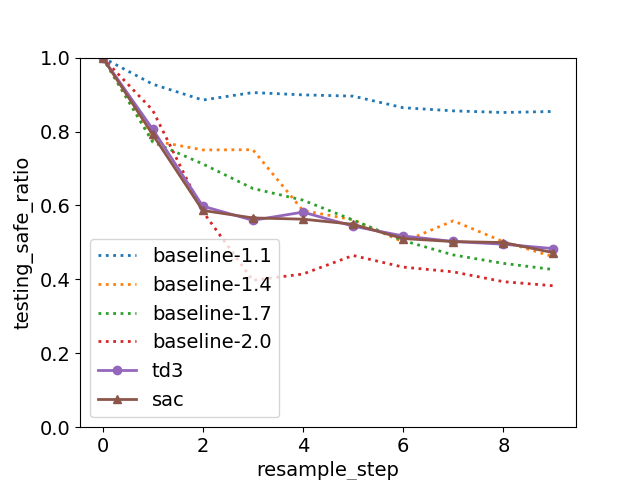}
        \caption{Testing safe ratio after RL agent training }
        \label{fig:lyapunov-gp-after}
    \end{subfigure}
    \setcounter{figure}{1}
    \captionof{figure}{Ratio of safe samples on Lyapunov NN compared with baseline $\alpha$}%
    \label{fig:lyapunov-gp}%
    \end{minipage}
    \setcounter{figure}{3}
    \begin{minipage}{\textwidth}
        \centering
        \begin{subfigure}[b]{0.45\textwidth}
            \centering
            \includegraphics[width=0.8\textwidth]{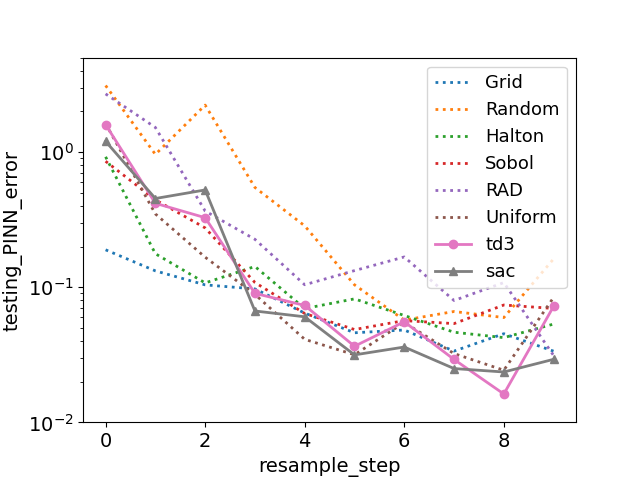}
            \caption{Before RL agent training}
            \label{fig:pinn-diffusion-test-before}
        \end{subfigure}
            \begin{subfigure}[b]{0.45\textwidth}
            \centering
            \includegraphics[width=0.8\textwidth]{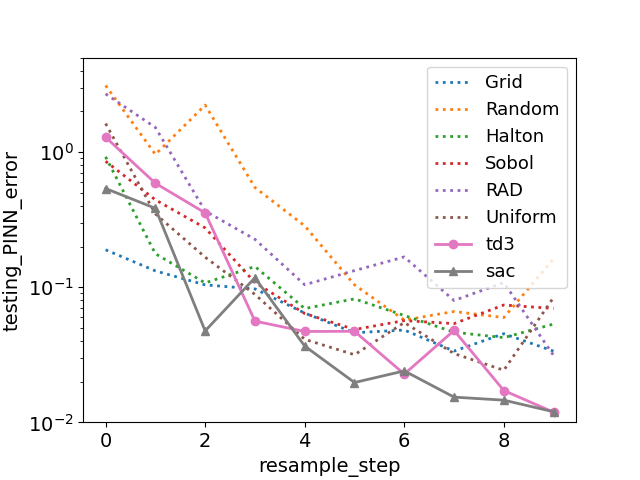}
            \caption{After RL agent training}
        \label{fig:pinn-diffusion-test-after}
        \end{subfigure}
        \setcounter{figure}{2}
        \captionof{figure}{Testing PINN error for PINN-Diffusion compared with baselines collocation selectors}%
        \label{fig:pinn_diffusion_test}%
    \end{minipage}
\end{figure*}
We train the Lyapunov neural network on inverted pendulum benchmark for 100 iterations
and update the safe level set (i.e., perform resampling) every 10 iterations. 
The RL agents are trained for 12,500 episodes,
using an action space corresponding to the expansion multiplier $\alpha$, bounded within the interval [1.1, 2.0].
During RL training, the pendulum’s pole length $l$ is randomized to encourage robustness, 
while evaluation is performed using a fixed pole length shared across all test runs and baseline methods.
As shown in Figure~\ref{fig:lyapunov-ssf-after}, both RL-based adaptive expansion policies outperform the fixed-multiplier baselines in terms of resulting safe set fraction. 
Figure~\ref{fig:lyapunov-ssf-step} further illustrates the progressive improvement throughout the RL training process.

To better understand why our agents outperform the baselines, 
we visualize the ratio of sampled points that are verified as safe before and after training the RL agent. 
As shown in Figure~\ref{fig:lyapunov-gp}, the untrained agents perform comparably to the median of the fixed-multiplier baselines. 
After training, however, the agents learn to expand more aggressively during the early stages,
allowing them to rapidly approach the boundary of the true safe set.
In later iterations, the policies become more conservative,
maintaining the ratio of safe samples close to 50\%, which supports stable expansion.
These results further highlight the effectiveness of our framework across a variety of learning settings.

\subsection{PINN-Diffusion}

\begin{figure}[t]
    \centering
    \includegraphics[width=0.4\textwidth]{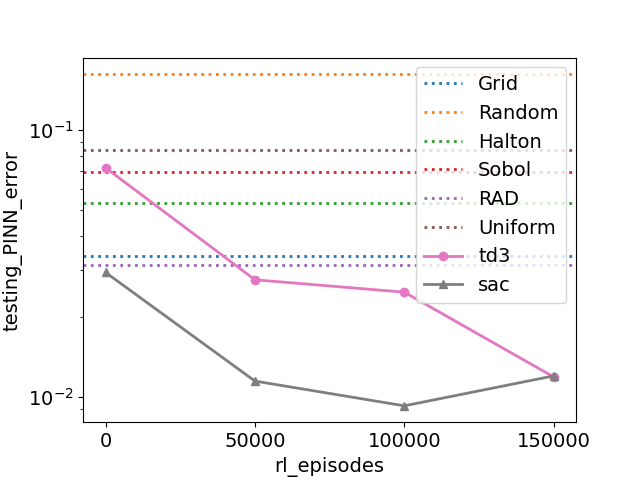}
    \captionof{figure}{Testing PINN error for PINN-Diffusion after 10,000 PINN iterations}
    \label{fig:pinn_diffusion_step10}
\end{figure}

We construct the first PINN environment based on the one-dimensional diffusion equation:
\begin{align}
    \frac{\partial u}{\partial t}=\frac{\partial^2 u}{\partial x^2}-(z^2\pi^2-1)e^{-t} \sin(z\pi x),\\ x\in[-1/z, 1/z], t\in[0,1]
\end{align}
where $z$ is a tunable constant used for environment randomization.	
The initial and boundary conditions are given by 
\begin{align}
    u(x,0) &= \sin(z\pi x),\\
    u(-1/z,t) &= u(1/z,t) = 0,
\end{align}
respectively.
This PDE admits an explicit solution: $u(x,t) = \sin(z\pi x)e^{-t}$.	
We compare our proposed framework against four random or quasi-random baselines, one manually defined adaptive sampling method, and a uniform mixture over all baselines.
For each collocation selection strategy, the PINN is trained using 50 residual points, which are resampled every 1,000 iterations over a total of 10,000 iterations.
The RL agent is trained for 150,000 episodes using different environment randomization parameters $z$, and all collocation selectors are evaluated using a shared test value of $z$.
As shown in Figure~\ref{fig:pinn-diffusion-test-before}, before RL training, 
the performance of the RL-based collocation selector is comparable to the uniform mixture of baselines.
After training, however, both RL agents (TD3 and SAC; Figure~\ref{fig:pinn-diffusion-test-after}) significantly outperform all baselines in terms of PINN error on testing collocations.
This demonstrates the effectiveness of our framework.
Additionally, Figure~\ref{fig:pinn_diffusion_step10} shows that after just 50,000 RL episodes, 
our agent already surpasses all baselines in terms of testing PINN error after 10,000 PINN iterations -
providing additional evidence of the proposed framework’s efficiency.

\begin{figure}[t]
    \centering
    \begin{subfigure}[b]{0.4\textwidth}
        \centering
        \includegraphics[width=\textwidth]{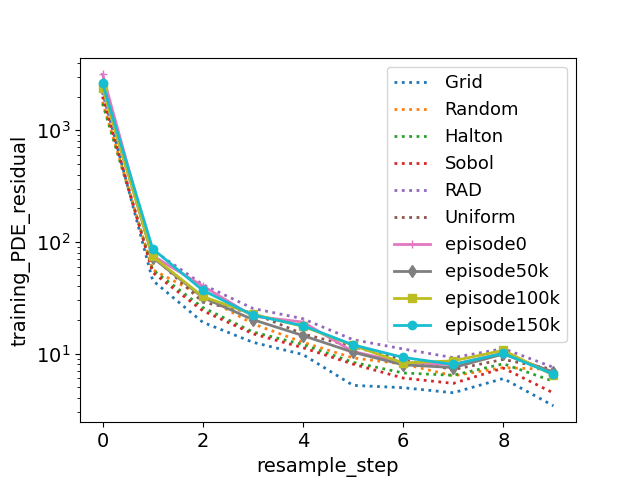}
        \caption{Training PDE residual}
        \label{fig:pinn-diffusion-pde}
    \end{subfigure}
    \begin{subfigure}[b]{0.4\textwidth}
        \centering
        \includegraphics[width=\textwidth]{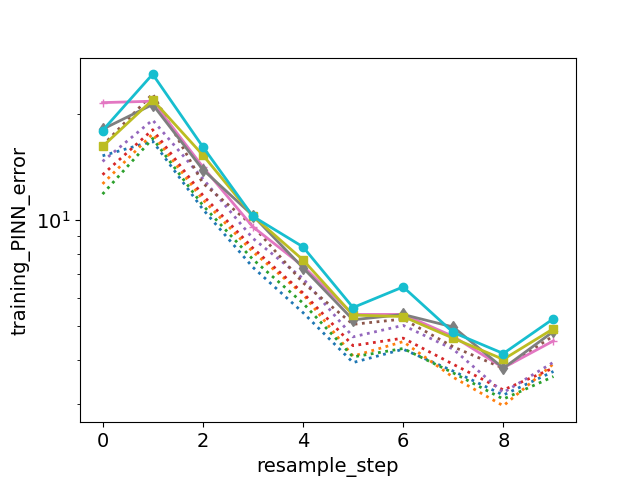}
        \caption{Training PINN error}
        \label{fig:pinn-diffusion-gt}
    \end{subfigure}
    \caption{Training PDE residual and PINN error from different collocation selectors.}
    \label{fig:pinn_diffusion_samePDE_td3}
\end{figure}
To further examine how the RL agent adapts to the environment,
we train the PINN using collocation sets selected by the TD3 agent after 150,000 RL episodes.
At each resampling step, we also generate collocation sets using baseline selectors,
as well as intermediate versions of the TD3 agent trained for fewer episodes.
For each collocation sets, we compute and plot both the PDE residual and the PINN error.
As shown in Figure~\ref{fig:pinn-diffusion-pde}, the RAD baseline tends to produce collocation points with the highest PDE residuals.
In contrast, Figure~\ref{fig:pinn-diffusion-gt} shows that the fully trained TD3 agent selects points that yield the higher PINN error,
highlighting how the proposed framework identifies challenging regions to improve training efficiency.

\subsection{PINN-Wave}
\begin{figure*}[t]
    \begin{minipage}{\textwidth}
        \centering
        \begin{subfigure}[b]{0.4\textwidth}
            \centering
            \includegraphics[width=\textwidth]{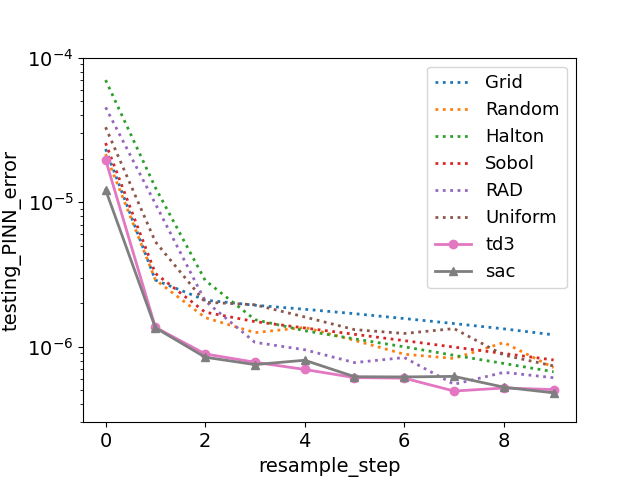}
            \caption{Testing PINN error after RL agent training}
            \label{fig:pinn-wave-test-after}
        \end{subfigure}
        \begin{subfigure}[b]{0.4\textwidth}
            \centering
            \includegraphics[width=\textwidth]{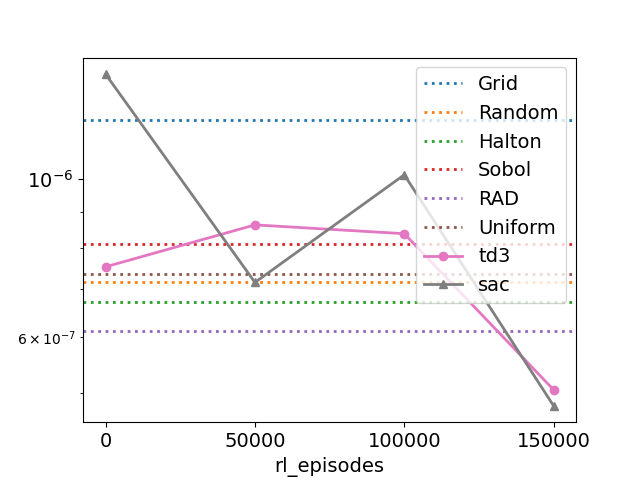}
            \caption{Testing PINN error at resampling step 10}
            \label{fig:pinn-wave-step}
        \end{subfigure}
        \setcounter{figure}{5}
        \captionof{figure}{Testing PINN error on PINN-Wave compared with baselines collocation selectors}%
        \label{fig:pinn_wave_test}%
    \end{minipage}
    \begin{minipage}{\textwidth}
        \centering
        \setcounter{figure}{7}
        \begin{subfigure}[b]{0.4\textwidth}
            \centering
            \includegraphics[width=\textwidth]{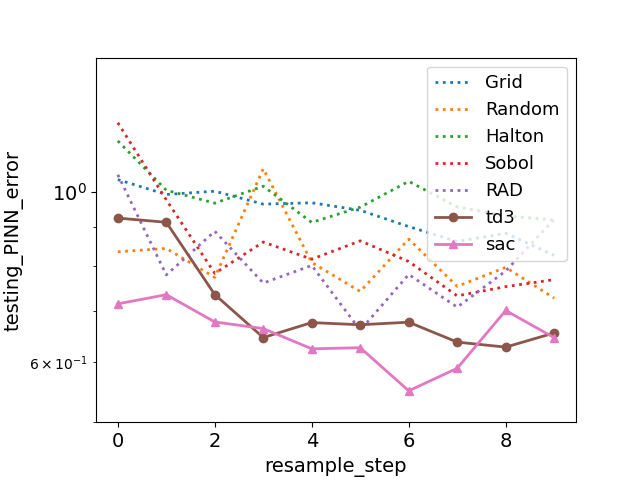}
            \caption{Testing PINN error after RL agent training}
            \label{fig:pinn-burgers-test-after}
        \end{subfigure}
        \begin{subfigure}[b]{0.4\textwidth}
            \centering
            \includegraphics[width=\textwidth]{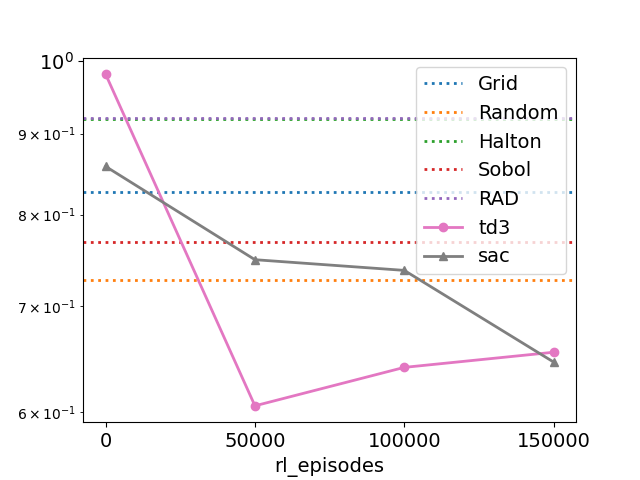}
            \caption{Testing PINN error at resampling step 10}
            \label{fig:pinn-burgers-step}
        \end{subfigure}
        \setcounter{figure}{6}
        \captionof{figure}{Testing PINN error on PINN-Burgers compared with baselines collocation selectors}%
        \label{fig:pinn_burgers_test}%
    \end{minipage}
\end{figure*}
We perform a similar experiment on the following one-dimensional wave equation:
\begin{align}
    \frac{\partial^2 u}{\partial t^2}-4z^2\frac{\partial^2 u}{\partial x^2}=0,\; x\in[0, 1],t\in[0,1],
\end{align}
where $z$ is an integer parameter used to randomize the environment.
The initial and boundary conditions are defined as:	
\begin{align}
u(x,0) &= \sin(\pi x)+\frac{1}{2}\sin(4\pi x),\\
\frac{\partial u}{\partial t}(x,0) &= 0,
u(0,t) = u(1, t) = 0,
\end{align}
respectively.
The exact solution to this PDE is given by:	
\begin{align}
    u(x,t) = \sin(\pi x)\cos(2z\pi t)+\frac{1}{2}\sin(4\pi x)\cos(8z\pi t).
\end{align}

For each collocation selection strategy, the PINN is trained using 50 residual points, 
resampled every 100 iterations over a total of 1,000 training steps.
The RL agents are trained for 150,000 episodes with varying environment randomization parameters $z$,
and all collocation selectors are evaluated using a shared test value of $z$.
As shown in Figure~\ref{fig:pinn-wave-test-after}, after 150,000 RL training episodes,
both TD3 and SAC agents significantly outperform all baselines. 
Additionally, we report the testing PINN error at resampling step 10 during RL training in Figure~\ref{fig:pinn-wave-step}, 
which shows a consistent improvement trend despite higher variance.
This further demonstrates the effectiveness of the proposed framework.	

\subsection{PINN-Burgers}

We perform a similar experiment on the following Burgers' equation:
\begin{align}
    \frac{\partial u}{\partial t}+u\frac{\partial u}{\partial x}=z\frac{\partial ^2 u}{\partial x^2},\; x\in[-1, 1],t\in[0,1],
\end{align}
where $z$ is a random viscosity parameter used to randomize the environment.
The initial and boundary conditions are defined as:	
\begin{align}
u(x,0) &= -\sin(\pi x),\\
u(-1,t) &= u(1, t) = 0,
\end{align}
respectively.
Unlike the other experiments, the Burgers’ equation does not admit a closed-form solution.
To address this, we first train a PINN for an extended number of steps to obtain a high-quality approximate solution, 
which is then used as a reference during RL training and evaluation.

For each collocation selection strategy, the PINN is trained using 50 residual points, resampled every 100 iterations over a total of 1,000 training steps.
The RL agents are trained for 150,000 episodes with varying environment randomization parameters $z$, and all collocation strategies are evaluated on a shared test value of $z$.

As shown in Figure~\ref{fig:pinn-burgers-test-after}, both TD3 and SAC agents outperform all baseline methods after 150,000 episodes.
We also report the PINN testing error at resampling step 10 during RL training in Figure~\ref{fig:pinn-burgers-step}, which illustrates the performance trajectory during training.

\section{Conclusion}
We presented a general framework for adaptive training input selection in neural networks governed by universal constraints.
Our approach formulates the input selection process as a Markov Decision Process,
allowing a learned policy to adjust in real time to evolving model states and training dynamics. 
By casting the problem as a sequential decision-making task,
the framework supports efficient, data-centric optimization in non-stationary settings.
Unlike prior methods based on static sampling rules or handcrafted heuristics,
our approach is model-agnostic.
We evaluate the framework on Lyapunov NNs and PINNs,
demonstrating consistent gains in convergence speed and solution quality over baselines.
These results highlight the benefits of adaptive data selection in constraint-driven learning and suggest opportunities for broader application and future optimization.

\newpage
\bibliographystyle{apalike}
\bibliography{references}


\clearpage
\appendix
\thispagestyle{empty}

\end{document}